\documentclass[10pt, journal]{IEEEtran}

\usepackage[english]{babel}
\usepackage{cite}
\usepackage[utf8]{inputenc} 
\usepackage[T1]{fontenc}    
\usepackage{hyperref}       
\usepackage{url}            
\usepackage{booktabs}       
\usepackage{amsfonts}       
\usepackage{nicefrac}       
\usepackage{microtype}      
\usepackage{xcolor}         
\usepackage{pifont}         

\hypersetup{
    colorlinks=true,
    linkcolor=blue,
    filecolor=magenta,      
    urlcolor=cyan,
    }
\usepackage{float}
\usepackage{graphicx}
\usepackage{prettyref}
\usepackage{amsfonts}
\usepackage{enumitem}
\usepackage{multirow}
\usepackage{mathtools}

\newcommand{\secrefprefix}{Section~}

\newcommand{\figrefprefix}{Figure~}
\newcommand{\tablerefprefix}{Table~}
\newcommand{\equationprefix}{Equation~}
\newcommand{\equationsprefix}{Equations~}

\newcommand{\cmark}{\ding{51}}%
\newcommand{\xmark}{\ding{55}}%

\newcommand{\intelndnschallenge}{Intel N-DNS Challenge}
\newcommand{\ndnschallenge}{Intel N-DNS Challenge}
\newcommand{\ndnsrepourl}{https://github.com/IntelLabs/IntelNeuromorphicDNSChallenge}
\newcommand{\ipgnet}{Intel DNS network}
\newcommand{\eg}{\it{e.g.}}

\newcommand{\ie}{\it{i.e.}}

\title{The Intel Neuromorphic DNS Challenge}

\author{
\IEEEauthorblockN{
    Jonathan Timcheck\IEEEauthorrefmark{1},
    Sumit Bam Shrestha\IEEEauthorrefmark{1},
    Daniel Ben Dayan Rubin\IEEEauthorrefmark{1},
    Adam Kupryjanow\IEEEauthorrefmark{2},
    Garrick Orchard\IEEEauthorrefmark{1}\IEEEauthorrefmark{3},
    Lukasz Pindor\IEEEauthorrefmark{2},
    Timothy Shea\IEEEauthorrefmark{1}, and
    Mike Davies\IEEEauthorrefmark{1}
    \\\IEEEauthorblockA{\small\IEEEauthorrefmark{1}Neuromorphic Computing Lab, Intel Labs}
    \\\IEEEauthorblockA{\IEEEauthorrefmark{2}Design Engineering Group Poland, Intel}
    \\\IEEEauthorblockA{\IEEEauthorrefmark{3}Work done during Intel employment}
    \\ndns@intel.com
}
\thanks{Manuscript submitted \today.}
}

\begin{document}

\maketitle

\begin{abstract}
A critical enabler for progress in neuromorphic computing research is the ability to transparently evaluate different neuromorphic solutions on important tasks and to compare them to state-of-the-art conventional solutions. 
The Intel Neuromorphic Deep Noise Suppression Challenge (\intelndnschallenge{}), inspired by the Microsoft DNS Challenge, tackles a ubiquitous and commercially relevant task: real-time audio denoising. 
Audio denoising is likely to reap the benefits of neuromorphic computing due to its low-bandwidth, temporal nature and its relevance for low-power devices.
The \intelndnschallenge{} consists of two tracks: a simulation-based algorithmic track to encourage algorithmic innovation, and a neuromorphic hardware (Loihi 2) track to rigorously evaluate solutions. For both tracks, we specify an evaluation methodology based on energy, latency, and resource consumption in addition to output audio quality.
We make the \intelndnschallenge{} dataset scripts and evaluation code freely accessible, 
encourage community participation with monetary prizes, and release a neuromorphic baseline solution which shows promising audio quality, high power efficiency, and low resource consumption when compared to Microsoft NsNet2 and a proprietary Intel denoising model used in production.
We hope the \intelndnschallenge{} will hasten innovation in neuromorphic algorithms research, especially in the area of training tools and methods for real-time signal processing. 
We expect the winners of the challenge will demonstrate that for problems like audio denoising, significant gains in power and resources can be realized on neuromorphic devices available today compared to conventional state-of-the-art solutions.
\end{abstract}

\section{Introduction}

\IEEEPARstart{N}{euromorphic} computing achieves excellent performance with power and latency savings for certain algorithms \cite{davies2021advancing}, and the field stands to greatly benefit from focusing on well-defined neuromorphic challenge problems motivated by recent progress. Challenge problems
facilitate the consistent evaluation and comparison of different approaches to solving important classes of problems and can help align researchers toward the most promising directions, thus accelerating progress. Historically, challenge problems have often spurred breakthroughs in the field of machine learning, {\eg}, MNIST \cite{lecun1998gradient}, CIFAR-10 \cite{krizhevsky2009learning}, and ImageNet \cite{deng2009imagenet}.
However, the less-mature field of neuromorphic computing lacks unifying challenge problems. Most results of benchmarking neuromorphic systems are bespoke, where custom tasks are conceived chiefly to highlight the capabilities of a given neuromorphic system, making it difficult to compare across different systems and solutions, whether neuromorphic or conventional \cite{tan2015benchmarking}. 

Any neuromorphic challenge problem must be chosen and structured carefully. 
A poorly-selected problem could direct focus in the wrong direction, on tasks for which neuromorphic hardware is unlikely to provide advantages over conventional hardware. This includes many existing popular machine learning tasks, such as those involving static image processing. 
Similarly, defining a challenge problem without an accompanying methodology for comprehensively evaluating neuromorphic compute cost makes it difficult to rigorously compare different solutions. 

Researchers have discussed at length what makes for good neuromorphic challenge problems and benchmarks \cite{davies2019benchmarks}, identifying qualities such as easy access and use, freely available data, not computationally prohibitive,  representative of an important real-world task, and unsaturated \cite{cramer2020heidelberg}. 
Existing neuromorphic benchmarks support these goals, but they are few in number and have key shortcomings. We briefly discuss several existing neuromorphic challenge problems in the following section.

\begin{figure}[t]
  \centering
  \includegraphics[width=0.9\columnwidth]{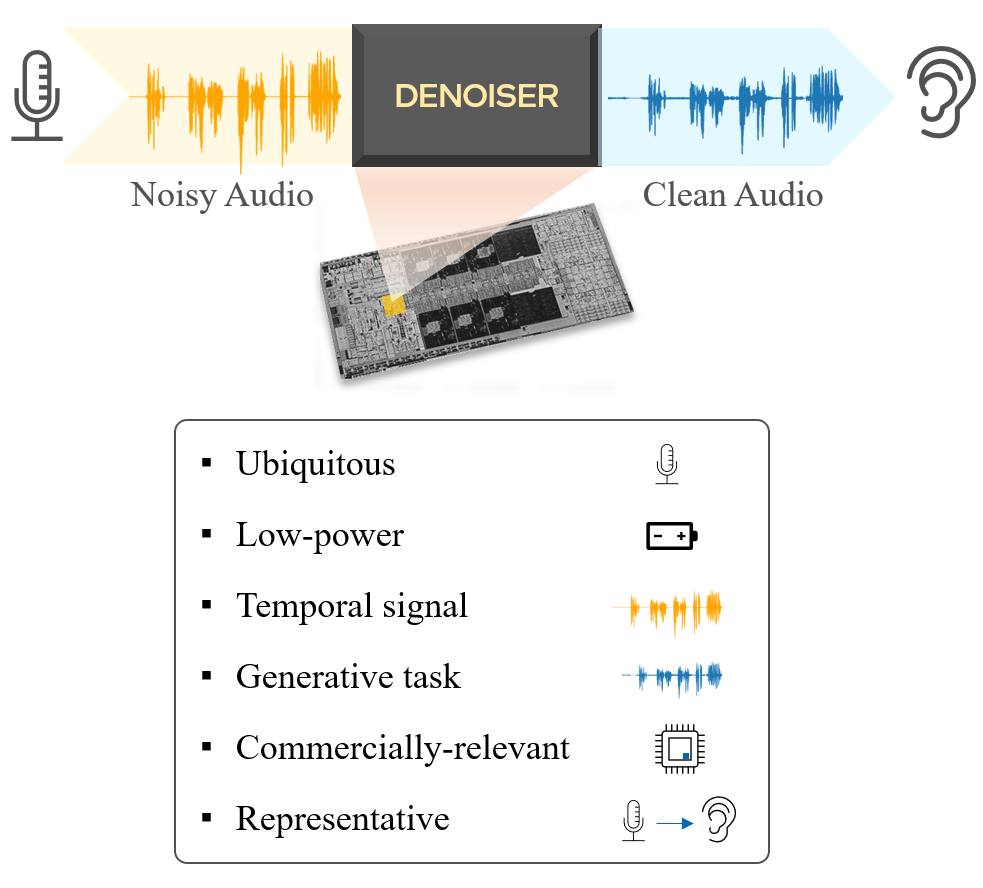}
  \caption{The audio denoising task. Audio denoising is ubiquitous and has many attributes that are likely to reap benefits from neuromorphic hardware. }
  \label{fig:audio-denoising-task}
\end{figure}

\subsection{Past neuromorphic challenge problems}\label{subsec:past-challenge-problems}

One of the first prominent neuromorphic challenge problems was image classification on the N-MNIST or N-Caltech101 datasets \cite{orchard2015converting}. N-MNIST and N-Caltech101 are neuromrophic versions of the classic  MNIST\cite{lecun1998gradient} and Caltech101\cite{FeiFei2004LearningGV} datasets: the neuromorphic datasets were captured using an event-based camera moving in a precise saccadic motion while pointed at a computer monitor displaying an MNIST or Caltech101 static image. 
While N-MNIST and N-Caltech101 were instrumental in advancing neuromorphic vision research and provided common datasets to compare various neuromorphic algorithms, the inherent source of information is a static image and lacks spatiotemporal information content\cite{iyer2021neuromorphic}, especially once the saccadic motion is compensated. 
Thus these datasets are generally not ideal for showcasing the full potential of neuromorphic computational models which aim to exploit neuronal dynamics inspired by biological neurons for efficient temporal signal processing
(\secrefprefix \ref{sec:neuromorphic-audio-denoising}).

Another popular neuromorphic vision challenge problem is gesture recognition on the DVS Gesture dataset \cite{amir17cvpr}. 
The DVS Gesture dataset is naturally matched to neuromorphic computing---the sparse, event-based, and spatiotemporal nature of dynamic vision sensor data naturally lends itself to neuromorphic processors that also possess these attributes.
Evaluated as a neuromorphic challenge problem, however, DVS Gesture uses specialized event-based sensor data which limits widespread applicability, and the dataset is small (1,342 instances). Neuromorphic solutions on DVS Gesture achieve a latency of 104ms\cite{amir17cvpr} on TrueNorth and 12.5ms on Loihi\cite{davies2021advancing} processing at 1ms per step. Further study shows that the accuracy on the task improves with a coarser timestep of up to 25ms\cite{yao2021temporal}. This indicates that the fine-grained temporal information in the DVS Gesture dataset may not be vital in this task; intuitively, common gestures are likely slow enough to be sufficiently captured on a slower timescale.

A popular neuromorphic audio challenge problem is keyword spotting on the Spiking Heidelberg Datasets \cite{cramer2020heidelberg}. The Spiking Heidelberg Datasets target the widely-applicable task of keyword spotting, and importantly, audio is pre-processed with a neuroscience-inspired cochlea model. 
This provides a consistent neuromorphic encoding to spikes upon which researchers can build their keyword spotting algorithms, thus facilitating simple and fair task performance comparisons across different spiking neural network (SNN) algorithms.
However, the cochlear encoding of the Spiking Heidelberg Datasets presents some critical shortcomings when viewed from the greater context of more general and more difficult audio processing tasks. Firstly, the information preservation of the cochlear encoding is unquantified, thus this encoding could artificially bottleneck the performance of keyword spotting, and perhaps severely bottleneck performance for more sophisticated audio processing tasks. Secondly, the power cost of computing the cochlear encoding is also unquantified, yet power is an important factor in real-world low-power audio processing systems. Indeed, how to encode an audio signal efficiently and faithfully for processing in a neuromorphic system is an open research question which plays an important role in our definition of the \intelndnschallenge.   

Other neuromorphic benchmarks have been proposed that target applications that are also well-matched to the spatiotemporal event-based neuromorphic computing style, such as Braille letter reading \cite{muller2022braille} and gesture recognition using electromyograph and dynamic vision sensor fusion \cite{ceolini2020hand}. 
However, these benchmarks involve niche sensors and applications, limiting their real-world impact and interest compared to more mainstream AI problems dealing with images, video, text, or audio.

\subsection{Audio denoising as a neuromorphic challenge}

In this work, we identify audio denoising as an excellent neuromorphic challenge task.
As detailed in subsequent sections, audio denoising has ubiquitous real-world applicability and plays to the strengths of neuromorphic computing.  
We have developed the \intelndnschallenge{} to make the task easily accessible, free to all, unsaturated, and designed specifically to make it easy to compare solutions over a comprehensive set of metrics.

The \intelndnschallenge{} is inspired by the Microsoft DNS Challenge, an audio denoising challenge that has been running since 2020 \cite{reddy2020interspeech, reddy2021icassp,reddy2021interspeech,
dubey2022icassp}. At a basic level, the Microsoft DNS Challenge has focused on improving speech denoising solutions as measured by human perceptual audio quality metrics and the Challenge included a track with the constraint that solutions must run in real-time on an Intel i5 or equivalent processor; essentially, the goal was to obtain the highest audio quality possible under the compute architecture constraint. In contrast, in the \intelndnschallenge{}, we are changing this architecture constraint and taking a more holistic approach to evaluating solutions by defining metrics for power and latency in addition to audio quality metrics.

The spirit of the \intelndnschallenge{} is to achieve production-level (near-SOTA) denoising performance in a system with at least an order of magnitude reduction in power, while also reducing latency, compared to real-time denoising solutions on conventional architectures. Our belief is that the neuromorphic computing features of Intel’s Loihi 2 chip---representative of future commercial neuromorphic devices---will enable the realization of such gains. Thus in the \intelndnschallenge{} we define one track focusing on evaluating solutions on existing neuromorphic hardware (Loihi 2) and another track focusing on neuromorphic algorithm development, which may motivate new features in future neuromorphic hardware.

The \intelndnschallenge{} has a 1-year timeline, but we invite the community to continue using the \ndnschallenge{} as a benchmark after the challenge ends.
More broadly, we view the N-DNS challenge as a single iteration in a continuing effort to develop challenge problems that help to advance neuromorphic computing to commercial maturity. 

We 
define the audio denoising task in \secrefprefix{}\ref{sec:audio-denoising-task},
discuss neuromorphic computing as it pertains to this work in \secrefprefix{}\ref{sec:neuromorphic-audio-denoising},
overview the Intel Neuromorphic DNS Challenge in \secrefprefix{}\ref{sec:n-dns-challenge}, describe the data in \secrefprefix{}\ref{sec:dataset}, specify evaluation criteria in \secrefprefix{}\ref{sec:evaluation}, %
 describe our baseline solution in \secrefprefix{}\ref{sec:baseline-solutions},
 address additional clarifications in \secrefprefix{}\ref{sec:clarifications}, 
 and summarize our contributions in \secrefprefix{}\ref{sec:conclusion}.
We make our code publicly available for obtaining the challenge data, evaluation pipeline, and the example baseline solution in the \href{\ndnsrepourl}{ \ndnschallenge{} Github  Repository ({\ndnsrepourl})} with a permissive MIT license.

\section{Primer on audio denoising}\label{sec:audio-denoising-task}

Digital audio signal denoising, also called audio signal enhancement, is a fast-growing research area, but its origin can be traced back to the late 70's and early 80's when Spectral Subtraction \cite{Boll_1979} and Wiener filter \cite{Ephraim_1985} algorithms were introduced. Subsequently, beamforming techniques were successfully adopted \cite{Higuchi_2016, Nakatani_2010}. While a significant advance, beamforming was not practical due to several limitations, namely, that multiple microphones are needed to perform noise reduction, Signal-to-Noise Ratio (SNR) improvement is highly correlated with the number of microphones, and compute complexity increases with the square of the number of microphones.
Furthermore, in the last few years, there has been an increased research interest in single microphone denoising. Single-microphone device configurations are omnipresent, and the utilization of Deep Neural Networks has enabled very successful single-microphone denoising  \cite{WaveNet_2017,luo2019conv,Yanxin_2020}. We address the single-microphone audio denoising task in the \intelndnschallenge{} (\figrefprefix{} \ref{fig:audio-denoising-task}).

Typically, the signal captured by a microphone contains a source signal, like speech or music, and stationary or non-stationary noises.
Stationary noises change amplitude and frequency profile slowly in time, whereas non-stationary noises vary quickly over time. Some examples of the former are an air conditioner, dishwasher, fan, or engine noises.
Examples of the latter are a baby crying, a dog barking, or keyboard typing. Notably, reduction of stationary noises is a significantly simpler task than the removal of non-stationary noises. 
Noise is an additive distortion defined in the time domain according to
\begin{gather}
 y(t) = x(t) + n(t),\label{additive_eq}
\end{gather}
where $x(t)$ is the amplitude of the source signal for time index $t$, $n(t)$ is the noise signal, and $y(t)$ is the noisy signal captured by the microphone. 

Furthermore, most recordings are conducted in a reverberant environment; {\eg}, in indoor conditions, the signal is contaminated by reverb. Noisy reverberant signals can be expressed as 
\begin{gather}
   y(t) = h(t) * x(t) + n(t),\label{combined_eq}
\end{gather}
where $h(t)$ represents impulse response and only a single noise source is represented.
Since reverb is a multiplicative distortion, most denoising algorithms will focus on noise removal \cite{Ochieng_2022}. There are alternative approaches that perform reverb reduction and noise removal in one shot \cite{Nakatani_2020} or use a cascade of processing with reverb reduction \cite{Nakatani_2010} in a first stage, followed by a denoising stage. Audio denoising refers specifically to the process of enhancing an audio signal by subtracting noise from it; this is the task in the \intelndnschallenge{} (\figrefprefix \ref{fig:audio-denoising-task}).

Audio denoising is commonly utilized in both real-time and non-real-time scenarios. An example of a real-time scenario is a voice call which is performed on an end-user device, such as a PC, phone, headset, or smart device, or inside applications like Microsoft Teams or Zoom.
In this use case, algorithms must not introduce latency greater than 40ms. Furthermore, the compute load must be light enough to fit into existing power and memory constraints without degrading user experience. 
Another example of a real-time application is speech enhancement in human-to-computer communication, where denoising is performed to improve the accuracy of downstream processing such as keyword spotting or Automatic Speech Recognition (ASR). 
There are other use cases, such as transcribing meeting minutes, where denoising can be performed offline. These are viewed as non-real-time scenarios that impose fewer restrictions on the algorithm, {\eg}, permitting non-causal filtering.

\subsection{Current state-of-the-art solutions}

Recently neural networks (NN) based algorithms have been extensively applied to audio denoising problems. Initial solutions focused exclusively on denoising quality and used large models to solve the problem with great breakthroughs in accuracy. However, as the models become more and more accurate, the focus has shifted to real-time denoising performance. In fact, the most recent Microsoft DNS Challenges have dropped the non-real-time track\cite{reddy2020interspeech, reddy2021interspeech, dubey2022icassp}.

Non-real-time solutions focus purely on the quality of denoising and are typically non-causal. Non-real-time solutions from the speech enhancement and source separation literature include attention architectures \cite{zhao2021monaural},
temporal convolutional networks (TCN) \cite{koyama2020exploring}, 
the Convolutional Time-domain audio separation Network (Conv TasNet) \cite{luo2019conv}, convolutional phase and amplitude processing (PHASEN) \cite{yin2020phasen}, and audio source separation with nested depthwise convolutional downsampling (SuDoRM-RF) \cite{tzinis2020sudo}. For the denoising task in speech enhancement, the desired enhancement is the removal of noise, and in source separation, the desired separation is between speech and noise.
Real-time solutions focus on making the network lightweight and causal while maintaining denoising performance. Some examples include causal forms of TCN, Conv TasNet \cite{koyama2020exploring}, and recurrent topologies with stacked LSTM or GRU \cite{braun2020data}.

The most common encoding-decoding method of choice is STFT-ISTFT\cite{braun2020data, yin2020phasen, zhao2021monaural} or its similar spectrogram transformation like DCT\cite{li2021real},  while methods like SuDoRM-RF\cite{tzinis2020sudo} directly process the raw audio samples. There are different approaches for processing the complex STFT input in the literature. Some methods only make use of the magnitude information\cite{braun2020data, koyama2020exploring, luo2019conv}, some process the magnitude and phase separately and combine them\cite{braun2022task}, while some process the complex spectrum directly using complex convolutional filters\cite{zhao2021monaural}.

The majority of the solutions use backpropagation-based supervised training. However, a wide variety of losses have been used in different works. The most common ones are the mean-square error of the resulting spectrum or maximization of the signal-to-noise ratio. A survey of various loss metrics used in audio denoising with their benefits is described in \cite{braun2021consolidated}. Some solutions even prioritize speech over suppression with an additional loss penalty term\cite{braun2022task}. In addition, unsupervised or semi-supervised training methods have also been investigated to achieve a general solution even on out-of-distribution datasets. A particularly interesting method is the teacher-student training method proposed in RemixIT\cite{tzinis2022remixit} where a teacher network trained on out-of-distribution data is used to bootstrap the noisy signals to multiply the variety of in-distribution data samples.

It is evident that noise suppression with deep neural networks is an active area of research with new methods being introduced regularly. Recent efforts have not only focused on the quality of denoising but also on the size of models and satisfying real-time requirements. There is a vast body of research from which to borrow for neuromorphic audio denoising.

\section{Neuromorphic audio denoising}\label{sec:neuromorphic-audio-denoising}

We chose the audio denoising task for this challenge because it presents an excellent opportunity for neuromorphic algorithm innovation (\figrefprefix \ref{fig:audio-denoising-task}). Audio denoising is a ubiquitous power-constrained task with commercial relevance.  It is often performed on mobile devices,  and every Intel Core\texttrademark{} CPU in production now includes AI hardware acceleration support for it. Given the significant compute load of today's denoising solutions, lowering the power with a neuromorphic solution could not only lead to longer battery lives and smaller form factors but could bring the functionality to even more power-constrained devices such as headsets, earbuds, hearing aids, and cochlear implants.
Moreover, it is a temporal signal processing task, which neuromorphic systems are expected to excel at \cite{davies2019benchmarks}.
Indeed, commercial neuromorphic vendors are already targeting speech-enhancing hearing aids, promising orders-of-magnitude gains \cite{femtosenseblog}.   
Looking forward, the audio denoising task represents a starting point for the development of more general neuromorphic audio processing algorithms that operate in real time with imperceptible latency, such as audio environment emulation, speech separation, voice morphing, and speech-to-speech language translation.

Furthermore, audio denoising is especially timely as a neuromorphic research vector. It is a generative task unsolved in neuromorphic computing, and audio is a low data-rate signal that is well-matched to current neuromorphic chips and designs that generally target low-power edge processing. 
Solutions can be readily compared to recent conventional machine learning advances, including models deployed in production, and can leverage insights, methods, and datasets from those recent efforts.

\subsection{Neuromorphic computing and Loihi 2}\label{sec:neuromorphic-computing}

Neuromorphic computing aims to apply fundamental principles of the brain's information processing mechanisms to engineered computing devices.  The brain consumes a mere 20 watts of power yet can execute remarkable feats of perception, planning, control, and learning while operating in real time processing sequential data streams. In contrast, our conventional computer systems today struggle to emulate even a narrow subset of such feats with much larger power budgets, even though they have the advantage of precisely engineered ultra-fast nanoscale transistors as a computational substrate \cite{mehonic2022brain}.
Indeed, biological inspiration is compelling. However, when computer architects go about designing neuromorphic systems, they face a fundamental question: What biology-inspired computational strategies unlock neuromorphic performance advantages versus conventional architectures?

Neuromorphic researchers have identified several promising strategies, such as analog computation, sparse connectivity, spike-based communication, in-memory computation, local synaptic learning rules, recurrent feedback, and stateful, dynamic neuron models \cite{schuman2022opportunities}. 
Subsets of these computational strategies are being implemented in hardware, {\eg}, novel analog devices\cite{wan2022compute}, 
analog computation in conventional circuits\cite{neckar2018braindrop, frenkel20180, schemmel2021accelerated, qiao2015reconfigurable}, 
digital processing with spike-based communication \cite{akopyan2015truenorth, davies2018loihi, pei2019towards, furber2014spinnaker, furber2020spinnaker}, 
and many others.

In the Intel Neuromorphic Computing Lab, we focus on designing all-digital neuromorphic processors that can be manufactured in state-of-the-art semiconductor process technology. The SOTA process enables direct comparisons to SOTA conventional architectures, and the all-digital character allows a broad range of architectural features to be rapidly prototyped with fully deterministic and repeatable execution. While the all-digital character sacrifices some efficiency benefits of analog computation, we believe it is most important to first rapidly explore the architecture-algorithm co-design space before undertaking the more difficult, slower, and currently less area-efficient path of analog circuit design and novel device engineering.  
We believe the subset of neuromorphic computational principles supported by our latest chip, Loihi 2, are sufficient to show significant gains in power and latency compared to conventional computer architectures, and that this will motivate further optimizations via more nascent neuromorphic computing principles.

Loihi 2 is a state-of-the-art neuromorphic chip designed to efficiently compute temporal dynamics in sparse networks using sparse, event-based communication \cite{loihi2techbrief}. Like its predecessor \cite{davies2018loihi}, Loihi 2 consists of neuron cores that compute the temporal dynamics of stateful neural models and a communication mesh optimized for spike-based communication. Loihi 2 implements a number of generalizations and optimizations motivated by the learnings and pain points of its predecessor. These include microcode-programmed neuron models, which enable a much wider variety of neurons as seen in the brain \cite{kandel2000principles} as well as in novel neuromorphic algorithms \cite{orchard2021efficient} and promising computational benefits in heterogeneous networks \cite{perez2021neural}. 
Loihi 2 also features graded spikes, {\ie}, spikes that carry an integer value, rather than binary spikes. While not biologically motivated, graded spikes are only marginally more costly to support than binary spikes in digital neuromorphic hardware and offer straightforward gains in algorithmic precision and processing speed. Loihi 2 also enhances Loihi 1's learning support so arbitrary local modulating factors (``third factors'') may be computed by postsynaptic neuron microcode.  We believe Loihi 2’s rich feature set is sufficient to unveil significant performance gains in tasks well-suited to temporal dynamics processing, hence the spirit of using Loihi 2 as a model for neuromorphic processing in the \intelndnschallenge{}.

\subsection{Neuromorphic audio processing and promising directions}\label{sec:past-and-promising}

The computational model implemented by neuromorphic processors such as Loihi 2 is that of a discretized dynamical system. 
Unlike conventional artificial neurons from machine learning, the state variables of a dynamical system evolve and process inputs in time---{\ie}, time is a fundamental ingredient of the computation. Thus we expect neuromorphic processors to naturally excel in temporal processing tasks, such as audio processing.
Indeed, precisely-timed spiking codes are well-known to underlie audio processing in the brain \cite{Kayser_Logothetis_Panzeri2010, Bialek_Wit1984, Martignoli_etAl2013}, and cochleas perform sophisticated transformations to encode incoming audio for effective processing \cite{Bialek_Wit1984, Martignoli_etAl2013}.
These insights from neuroscience provide clear hope for the feasibility and success of neuromorphic audio processing, and recent progress on tasks such as keyword spotting provide some evidence thereof\cite{orchard2021efficient, cramer2020heidelberg, Anumula18fns, YargaRouatWood2022}.

One can immediately ascertain three critical research questions when designing a neuromorphic audio processing system: (1) How to efficiently represent an audio waveform with high fidelity in the neuromorphic domain? (2) How to efficiently perform the desired audio processing (denoising) on this neuromorphic representation? and (3) How to efficiently invert the neuromorphic representation to yield an output (waveform)?

A natural place to start answering these questions is to start with the first: how to efficiently represent a waveform in the neuromorphic domain.
There exist a variety of possibilities for representing data neuromorphicly---{\eg}, binary spikes, graded spikes, population codes, sparse distributed codes, and phase codes---and a variety of encoding algorithms---{\eg}, biology-inspired cochleogram models \cite{orchard2021efficient, cramer2020heidelberg}, Short-Time Fourier Transforms (STFT)  \cite{grochenig2001foundations}, and  Mel-frequency cepstral coefficients (MFCCs) \cite{rabiner1993fundamentals}.
Taking inspiration from biology, in developing our baseline solution for the \intelndnschallenge{}, we initiated our study of neuromorphic audio encodings on cochleogram models, which can provide sparse representations in binary spikes, high sensitivity, frequency selectivity, large dynamic range, pitch-shifting, and self-peak normalization \cite{Magnasco2003, Hudspeth_etAl2010, Martignoli_etAl2013, orchard2021efficient}.
However, we quickly realized that cochleogram models such as \cite{zilany2014updated, cramer2020heidelberg} are generally computationally expensive to invert with high fidelity, which is prohibitive for a low-power denoising system.
As an alternative, we developed our initial baseline solution for the \intelndnschallenge{} using a more conventional audio encoding, the Short-time Fourier Transform (STFT) \cite{grochenig2001foundations}, which is easy to invert and has perfect fidelity (aside from quantization and numerical error); furthermore, the STFT encoding can take advantage of graded spikes which are supported on Loihi 2.

While we select an STFT encoding for our baseline, we emphasize that new solutions to the \intelndnschallenge{} have a wide range of encoding strategies to explore, {\eg}, designing invertible bio-inspired cochleogram models, utilizing sparse STFTs \cite{orchard2021efficient}, or even encoding schemes that depend on feedback from other portions of the neuromorphic denoising system, much like the recurrent feedback connections from deeper areas of the brain to more low-level sensory encoding areas. Importantly, the encoding used in a neuromorphic audio processing system must be co-designed with the task for efficient operation; indeed, such synergistic design is observed in biology \cite{DeWeese_Wehr_Zador2003, SmithLewicki2005}.

Secondly, after audio is encoded, the actual execution of the audio processing in the neuromorphic domain is a very open research opportunity. 
Neuromorphic audio processing systems can employ a wide variety of strategies to perform processing in the neuromorphic domain, such as simplistic DNN conversion \cite{blouw2019benchmarking}, using a network of feedforward or recurrent leaky integrate-and-fire neurons \cite{yin2020effective, shrestha2022spikemax, cramer2020heidelberg}, a network of complex resonate-and-fire neurons \cite{orchard2021efficient}, or a sigma-delta neural network as we describe in the following subsection for our baseline solution.
Methodologies inspired by conventional deep learning, {\eg}, multi-timescale networks\cite{tzinis2020sudo, tzinis2022remixit, koyama2020exploring} or attention\cite{zhao2021monaural}, if mapped efficiently to the neuromorphic domain, could be promising directions as well.  
And finally for completeness---to address the third question posed above---decoding the output of the neuromorphically-processed audio again depends on the processing used and must be tailored appropriately to operate in an efficient manner.

Thus we see much opportunity for innovation throughout a neuromorphic processing pipeline---encoding, processing, and decoding. Furthermore, the audio denoising task represents just one potential audio processing task that opens the door to tackling many others with methods that are transferable to other signal processing domains such as wireless, biosensors, and control.

\subsection{Baseline neuromorphic solution}

We have developed a simple baseline neuromorphic solution to the audio denoising task, and we already begin to see evidence of significant energy efficiency gains from using neuromorphic features. The baseline solution uses a sigma-delta neural network (SDNN), an adaptation of a conventional feedforward ReLU neural network architecture that exploits sparse message passing with graded spikes and stateful neurons---computational strategies that can be implemented efficiently in neuromorphic architectures and that are supported by Loihi 2 in particular. The SDNN baseline solution achieves similar audio quality to a conventional baseline solution NsNet2 from the Micrsoft DNS Challenge 2022, but with an order of magnitude fewer operations and less than half its latency. We provide a more detailed overview of the baseline solution architecture and its performance in \secrefprefix{} \ref{sec:baseline-solutions}.

Importantly, our SDNN baseline solution is a very basic feedforward architecture, and does not exploit several of the aforementioned neuromorphic features that perform well on Loihi 2 (\tablerefprefix \ref{table:neuromorphic-features-used-by-baseline}). 
As new solutions incorporate more of these features, such as recurrent and sparse connectivity, we anticipate further significant improvements in power and model size.

\begin{table}[H]
\centering
\caption{Neuromorphic features that are performant on Loihi 2 and their utilization in our N-DNS baseline solution.}
\begin{tabular}{ p{5cm} c}
 Neuromorphic feature           & In baseline solution \\ \hline
 Sparse activity                & \cmark \\
 Sparse connectivity            & \xmark \\
 Recurrence                     & \xmark \\
 Stateful neurons               & \cmark \\
 Neuron temporal dynamics       & \xmark \\
 Synaptic plasticity            & \xmark \\
 Graded spikes                  & \cmark \\
 Delay as computational element & \cmark \\
 \hline
\end{tabular}\label{table:neuromorphic-features-used-by-baseline}
\end{table}

\section{Intel Neuromorphic DNS Challenge}\label{sec:n-dns-challenge}

Just like the Microsoft DNS Challenge, 
The objective of the Intel Neuromorphic DNS Challenge 
is to create a system that removes the noise from noisy human speech in real-time. However, in contrast to the denoising system that runs on a conventional CPU in the Microsoft DNS Challenge, the \intelndnschallenge{} targets the Loihi 2 neuromorphic processor aiming to  realize the neuromorphic system’s potential for power and latency improvements. To this end, the \intelndnschallenge{} hosts two tracks:
\begin{enumerate}
    \item Algorithmic. The objective in Track 1 is to develop a high-quality audio denoising solution that operates efficiently on a neuromorphic system. The algorithm is not required to run on actual neuromorphic hardware, but rather will be simulated on conventional hardware. Latency and a neuromorphic proxy power are estimated. 
    \item Loihi 2. The objective in Track 2 is to develop a high-quality audio denoising system that operates efficiently on Loihi 2 \cite{loihi2techbrief}. The power and latency of the denoising solution will be measured by running it on actual Loihi 2 hardware.

\end{enumerate}
Track 1 provides freedom to explore a wide range of neuromorphic denoising solutions, without the need to demonstrate the solutions on actual neuromorphic hardware; this track is intended for rapid development and potentially to inspire future neuromorphic hardware features. 
Track 2 guarantees that neuromorphic denoising solutions can indeed run on actual neuromorphic hardware. This track provides a rigorous demonstration of power and latency benefits realized by neuromorphic hardware. 

\begin{figure*}[!ht]
  \centering
  \includegraphics[width=0.8\textwidth]{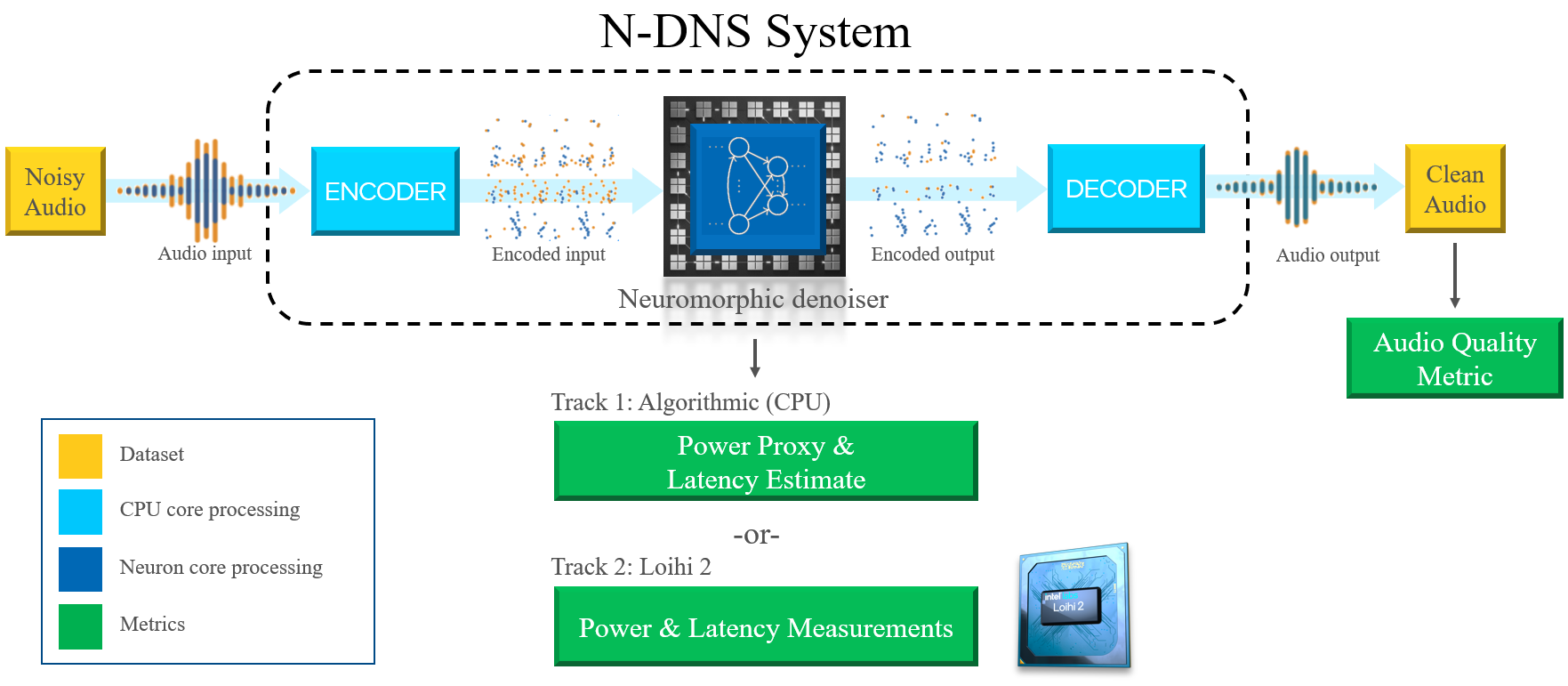}
  \caption{Intel Neuromorphic DNS Challenge Solution Structure. 
  Input noisy audio is encoded before it enters the neuromorphic denoiser (N-DNS). The neuromorophic denoiser processes its input, and the output of the neuromorphic denoiser is decoded to produce the final output clean audio. The encoder, decoder, and neuromorphic denoiser are the constituents of a solution to the \ndnschallenge{} and their power and latency are evaluated, in addition to the output audio quality. In Track 1, all components run on CPU, while in Track 2, the neuromorphic denoiser runs on Loihi 2.
  }
  \label{fig:solution_structure}
\end{figure*}

Both tracks follow the same structure: noisy audio is encoded into a form suitable for processing on a neuromorphic system, processed on a neuromorphic system (simulated for Track 1, or real hardware system for Track 2), and decoded into a clean output audio waveform (Figure \ref{fig:solution_structure}). Solutions are evaluated by an audio quality metric and a computational resource usage metric and are subject to a minimum audio quality and maximum latency (real-time) requirement.

The selection procedure for the winner of each track is described in the \href{\ndnsrepourl}{\intelndnschallenge{} Github Repository}, along with challenge logistics and timeline.
Solutions will be judged not only on the measured or estimated computational metrics, but also on commercial relevance, broader research impact, and quality of solution write-up.
We describe the dataset, evaluation metrics, 
and an example baseline solution in the following sections. 

\section{Dataset}\label{sec:dataset}
The Intel N-DNS dataset is derived from the Microsoft DNS Challenge dataset, which is a corpus of human speech audio samples of various categories including but not limited to English, German, French, Spanish, Russian and various categories of noises (\href{https://github.com/microsoft/DNS-Challenge}{ DNS Challenge Github Repository}). We provide a synthesizer script that generates 30-second segments of clean (ground truth), noise (additive), and noisy (ground truth + noise) audio data for both the training and validation dataset in the challenge repository. For training the network, participants are free to choose and/or tweak the data synthesis parameters or choose only a subset of the Microsoft DNS Dataset language and noise categories, or even include additional speech and noise corpus for synthesis. The default is 500 total hours (60,000 samples) of audio data with the synthesized SNR between 20dB to -5dB at 16kHz with a bit depth of 16 bits. 
The validation set, on the other hand, is generated using the default settings in the audio synthesis script.

The testing data for \intelndnschallenge will be provided at a later point after participant models are frozen. Thereafter, there can be no changes to the submitted models in order to ensure a fair evaluation on the test set. 
The characteristics of the testing data will be similar to the training and validation set. 
Note that this model freeze is only a feature for administering the challenge in a fixed timeline with blinded test set, and we encourage the continued use of the \intelndnschallenge{} resources and framework as a general, non-time-bound challenge problem for neuromorphic research.

In addition, we include general dataloader modules in the \intelndnschallenge\ that load the clean, noise, and noisy audio from the training, validation, and testing samples. Optionally, the dataloader also provides metadata about synthesized audio samples like the clean audio sources, noise sources, the noise mixture level and so on.

\section{Evaluation} \label{sec:evaluation}

There is no single metric that captures the overall performance of a solution in the \intelndnschallenge. Instead, there are multiple metrics that characterize different dimensions of performance. Naturally, we must quantify the output audio quality of the N-DNS system, and so we define metrics for this related to signal-to-noise ratio and perceptual audio quality. Equally important for the objective of the challenge is to assess computational resource costs: latency to ensure real-time processing, power to quantify energy efficiency, and chip resources required to support the solution on neuromorphic hardware.
With these four performance dimensions covered, we can comprehensively evaluate each solution. We also consider certain derived figures of merit, such as power-delay product, a common quantity used to represent the tradeoff between speed and energy efficiency in electronics systems. 

This collection of metrics allows us to compare solutions designed for different points in performance space, {\ie}, its positioning on a Pareto frontier with top-performing solutions designed for low-power or high-power, with correspondingly lower or higher audio quality. 

\subsection{Audio quality metrics and minimum audio quality improvement}

\subsubsection{SI-SNR metric}
Task performance in the N-DNS Challenge is measured as the output audio quality; we use the Scale-Invariant Source-to-Noise Ratio (SI-SNR)---SI-SNR is a common metric in the audio processing literature ({\eg}, \cite{bahmaninezhad2019comprehensive, le2019sdr}).
SI-SNR measures how clear the human speech is above the noise in the output of the N-DNS system, similar to a Source-to-Noise Ratio (SNR) \cite{le2019sdr}.
But importantly, SI-SNR is also scale-invariant---{\ie}, changing the overall magnitude (volume) of the output does not change the SI-SNR; intuitively, we do not wish to favor solutions over others' that simply increase the output volume. 

For a single input waveform, a real-valued 
zero-mean
 vector $s$, and the corresponding output waveform from the N-DNS system $\hat{s}$, the SI-SNR is defined as 
\begin{equation}
\text{SI-SNR} := 10 \log_{10} \frac{|| s_\text{target} ||^2 }{||e_\text{noise} ||^2},\label{eq:si-snr}
\end{equation}
where
$
s_\text{target} := \frac{\langle \hat{s}, s \rangle s}{||s||^2} 
$ and $
e_\text{noise} := \hat{s} - s_\text{target}
$.

We choose SI-SNR as one of our metrics for its simplicity and generality, rather than more complicated audio quality metrics, such as speech-to-text word accuracy used in the Microsoft DNS Challenge \cite{dubey2022icassp}. 
The focus of the N-DNS challenge is on neuromorphic algorithm innovation; this in itself constitutes a sufficiently challenging task. Moreover, we view the audio denoising task as a representative of a general audio processing workload, and some commercial applications may not specifically prioritize human-listener perceptual quality. Finally, the SI-SNR can be conveniently used as a loss function for machine learning approaches.

The mean $\text{SI-SNR}$ on the test set will be used to compare solutions. A script for computing mean SI-SNR is provided in the \href{\ndnsrepourl}{\intelndnschallenge{} Github Repository}.

\subsubsection{Minimum SI-SNR improvement}

Since solutions in the \ndnschallenge{} are evaluated holistically, solutions may target high audio quality by using a large amount of power, or lower audio quality using a smaller amount of power, or any audio quality-power point in between.
However, to ensure that the audio denoising task is being solved to some significant extent, we require solutions to achieve a minimum audio quality improvement over the noisy input audio quailty. 
Moreover, per our emphasis on neuromorphic computing, we require that the neuromorphic component of the N-DNS system be responsible for a significant portion of the audio quality improvement; a solution may optionally perform some denoising in the encoder and decoder, but the spirit of the \ndnschallenge{} is in performing neuromorphic denoising. 

Therefore, we define two measures of audio quality (SI-SNR) improvement (i) relative to 
(1) the noisy data ($\text{SI-SNRi}_{\text{data}}$) and 
(2) the encode+decode processing ($\text{SI-SNRi}_{\text{encode+decode}}$),
expressed by the following inequalities:
\begin{align}
\text{SI-SNRi}_{\text{data}} &> 3\text{dB}\label{eq:min-si-snr-system} \\
\text{SI-SNRi}_{\text{enc+dec}} &> 3\text{dB}, \label{eq:min-si-snr-neuromorphic}
\end{align}
where
\begin{itemize}
    \item $\text{SI-SNRi}_\text{data} = \text{SI-SNR}_{\text{full system}} - \text{SI-SNR}_{\text{data}}$,
    \item $\text{SI-SNRi}_\text{enc+dec} = \text{SI-SNR}_{\text{full system}} - \text{SI-SNR}_{\text{enc+dec}}$,
    \item $\text{SI-SNR}_{\text{full system}}$ is the mean test-set SI-SNR from the full N-DNS system (input audio $\rightarrow$ encode $\rightarrow$ neuromorphic denoiser $\rightarrow$ decode $\rightarrow$ output audio),
    \item $\text{SI-SNR}_{\text{enc+dec}}$ 
    is the mean test-set SI-SNR from running only encoder and decoder
    (input audio $\rightarrow$ 
    encode $\rightarrow$
    decode $\rightarrow$
    output audio), and
    \item $\text{SI-SNR}_{\text{data}}$ 
    is the mean test-set SI-SNR on the noisy input audio (no transformations).
\end{itemize}

{\equationprefix} \eqref{eq:min-si-snr-system} ensures that the solution achieves a minimum audio quality improvement, and {\equationprefix} \eqref{eq:min-si-snr-neuromorphic} ensures that the neuromorphic denoiser itself is responsible for a minimum audio quality improvement. These definitions allow for some amount of denoising to occur in the encoder and decoder, but critically, adding the neuromrophic denoiser must further improve audio quality. Similarly, additional pre/post-processing could be performed within the neuromorphic denoiser itself, to reduce the amount of computation in the encoder and decoder. But importantly, the computations allocated to the encoder/decoder or the neuromorphic denoiser are accounted for differently in the computational resource and chip usage metrics, as described in later sections.

\subsubsection{DNSMOS metric}

For audio signals, the perceptual quality of the audio signal is important in addition to the signal quality measured by SI-SNR. We use the widely adopted DNSMOS\cite{reddy2022dnsmos} metric to evaluate the perceptual quality of the solution. In DNSMOS, the perceptual quality score is predicted by a deep network that is trained to reflect the human perceptual quality expressed in Mean Opinion Score (MOS) in its training corpus. MOS score ranges from 1 to 5, where 1 corresponds to poor quality, and 5 corresponds to excellent quality. DNSMOS is particularly effective because it has been shown to generate scores that are highly correlated with human perceptual assessment \cite{reddy2022dnsmos} compared to other similar methods like Perceptual Evaluation of Speech Quality (PESQ) \cite{rix2001perceptual}, Perceptual Objective Listening Quality Analysis (POLQA) \cite{Beerends2013PerceptualOL}, or VisQL \cite{ViSQOL_2015}. There exist commercial alternatives like 3QUEST, but its use is limited due to its proprietary nature. 

A DNSMOS score consists of three values: speech signal quality~(SIG), background noise quality~(BAK), and overall audio quality~(OVRL). From the perspective of speech enhancement, the SIG score reflects the change in speech quality due to processing. Usually, most denoising algorithms do not improve SIG score significantly compared to the unprocessed signal. BAK score reflects the degree of noise present in the signal. Thus, after a speech enhancement, a significant improvement in this score is expected. Finally, OVRL score reflects the general audio quality assessment. It is not a simple average of SIG and BAK scores, but rather a general assessment of audio quality. After denoising, a signal should have a higher OVRL score.

DNSMOS provides a valuable additional facet in evaluating audio quality in the \intelndnschallenge{}. In addition, it gives another point of comparison to existing denoising systems; namely, DNSMOS (OVRL) was used in the Microsoft DNS Challenge \cite{dubey2022icassp}. However, we note that while DNSMOS is an important metric, we emphasize that it is not the only metric used for the evaluation of audio quality in the \intelndnschallenge{}; indeed, the spirit of the \intelndnschallenge{} is directed toward holistic innovation on neuromorphic denoising systems. Furthermore, to minimize the complexity of the \intelndnschallenge{}, we choose to not introduce additional audio quality metrics, such as STOI \cite{taal2011algorithm}, as the pairing of SI-SNR and DNSMOS already provides an objective and a perceptual audio quality evaluation.

\subsection{Computational resource usage and real-time requirement}\label{subsec:comp-resource-and-real-time-req}

Computational resource cost is evaluated in terms of power, latency, number of parameters, and model size.  To qualify as a real-time solution, the end-to-end latency must not be greater than 40ms. We measure power and latency on neuromorphic hardware in Track 2, but for Track 1, we introduce proxy metrics.  

\subsubsection{Latency}

An audio denoising system takes some amount of time to process input audio as the audio streams into the system; this results in the output human speech being delayed relative to the input human speech. This delay is the latency of the denoising system. For the denoising system to be considered real-time, the latency must be less than some human perceptual threshold, which in our case we choose to be 40ms.

We define latency as the maximum time difference between any corresponding segment of audio in the input and output of the N-DNS system. Intuitively, the longest delay in any segment of audio is the overall delay the output must be presented at in order to not introduce playback speed fluctuations in the output audio. 

\begin{figure*}[!ht]
    \centering
    \includegraphics[width=0.8\textwidth]{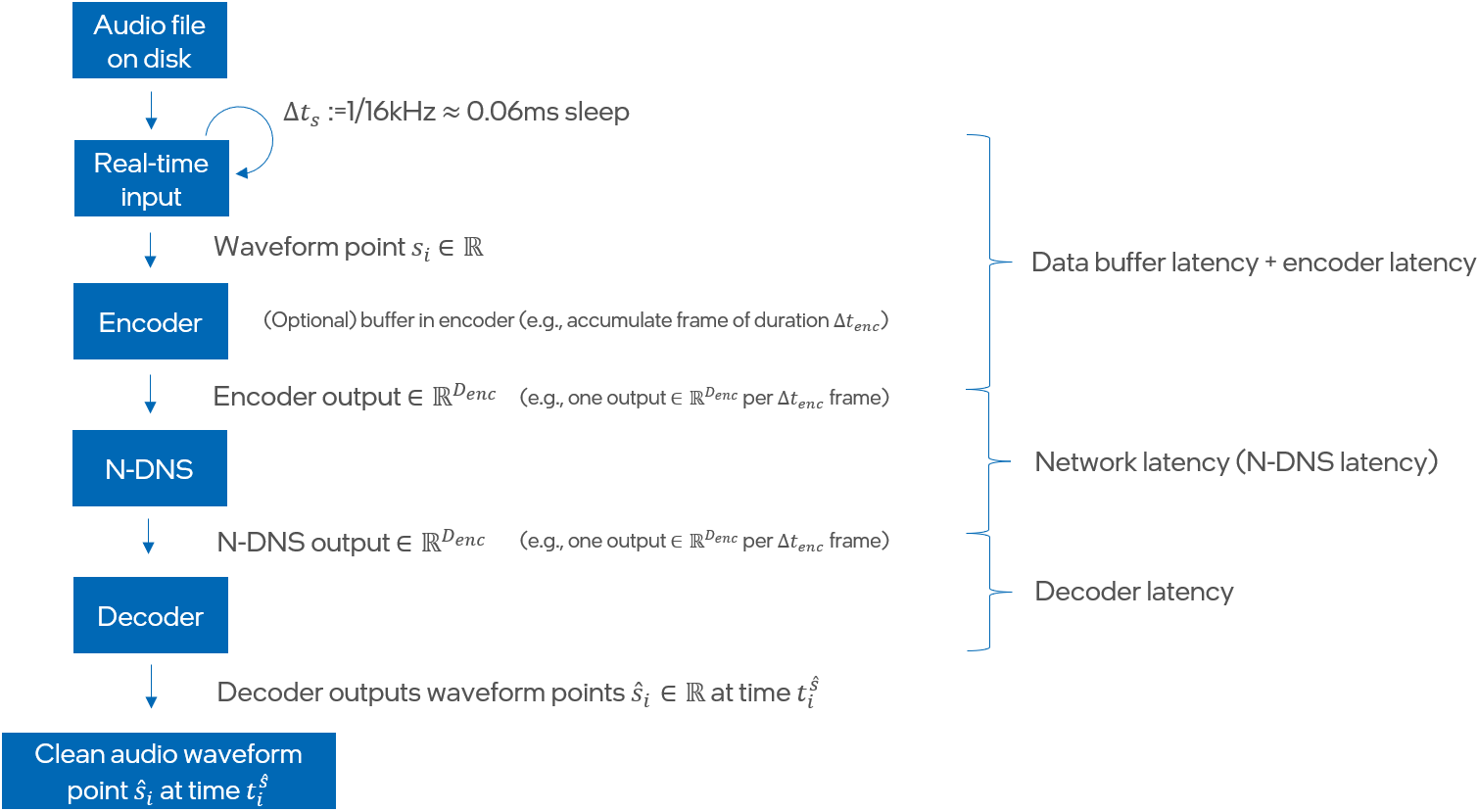}
    \caption{Real-time N-DNS pipeline. Latency is calculated by considering a real-time input propagating through an entire N-DNS system. This includes latency from buffering the input data, latency from CPU processing time of the encoder and decoder, and any latency introduced by the N-DNS component ({\eg}, a network that was trained to output audio delayed relative to the input). 
    }
    \label{fig:latency}
\end{figure*}

Latency should be calculated by considering a real-time input propagating through an entire N-DNS system (\figrefprefix \ref{fig:latency}). This includes \emph{data buffer latency}, \emph{encoder-decoder latency}, and \emph{network latency (N-DNS latency)}:
\begin{enumerate}
    \item \emph{Data buffer latency} is the time required to collect the audio stream to process one discrete timestep, however that may be defined for a given encoding scheme. For the STFT encoder in our SDNN baseline solution, the data buffer latency is equal to the STFT {\it window length}.
    \item \emph{Encoder-decoder latency} is the wall clock processing time to encode one discrete timestep-worth of the audio data, to be processed by the N-DNS network, and decode it back.
    \item \emph{Network latency (N-DNS latency)} is the latency introduced by the neuromorphic denoising network. It is measured by the maximum cross-correlation between the clean target audio and the denoised audio from the  network.
\end{enumerate}

In Track 1, notably, the (CPU) processing time for the neuromorphic denoiser (N-DNS) portion of the solution is not included in the latency calculation. We assume that the neuromorphic processing time will be small relative to the real-time timestep due to the high degree of parallelization in neuromorphic algorithms and hardware. 
In the case of the baseline SDNN, for example, the network must process a new STFT frame every 8 milliseconds, whereas Loihi 2 circuits typically complete all spike processing and neuron evaluations for a timestep within microseconds.
We provide an example Track 1 latency calculation in \secrefprefix{} \ref{sec:baseline-solutions}.

For Track 2, latency is simply measured on a reference CPU + Loihi 2 system. The measurement methodology and an example will be provided in the \href{\ndnsrepourl}{\intelndnschallenge{} Github Repository} later in the challenge.

\subsubsection{Power}

For Track 1, we calculate a power proxy by estimating the effective number of synaptic operations per second:
\begin{equation}
P_\text{proxy} = \text{Effective SynOPS} = \text{SynOPS} + 10 \times \text{NeuronOPS}, \label{eq:power-proxy}
\end{equation}
where SynOPS and NeuronOPS are the mean number of synaptic operations and mean number of neuron updates, respectively, per second of audio processed in the N-DNS stage. Synaptic operations and neuron operations can be considered the computational primitives of a neuromorphic system, and energy usage is roughly proportional to their number, with the approximate weighting of the energy of one neuron operation being equal to that of about 10 synaptic operations in our experience with the Loihi architecture \cite{davies2018loihi}. 
While $P_{\text{proxy}}$ gives only a crude power estimate, it provides a simple and sufficiently reliable assessment of a neuromorphic power advantage without needing to run on neuromorphic hardware.

The power consumption of the encoder and decoder is not taken into account in Track 1. We make this choice for simplicity, in expectation of the neuromorphic power dominating in realistic solutions. Note that the real-time requirement implicitly bounds the amount of computation that can be performed in the encoder and decoder. 

In Track 2, the encoder and decoder are implemented on a CPU and the N-DNS stage is implemented on a Loihi 2 system. The power is simply measured on a reference CPU and Loihi 2 system. 
Note that since both CPU and Loihi 2 power components will be measured, any attempt to implement a disproportionate amount of the denoising functionality inside the encoding/decoding CPU stages will result in a very high power result.
Details for measuring power on a reference system will be provided in the \href{\ndnsrepourl}{\intelndnschallenge{} Github Repository}.

\subsubsection{Power delay product}
The Power Delay Product~(PDP) metric combines both latency and power efficiency in one number that allows comparing between different solutions that make different tradeoffs between running faster at higher power versus running slower at lower power. For Track 1, a proxy PDP measure is given by
\begin{equation}
    PDP_{\text{proxy}} = P_{\text{proxy}} \times L,
    \label{eq:pdp_proxy}
\end{equation}
which is in units of Ops because $P_{\text{proxy}}$ (\equationprefix \eqref{eq:power-proxy}) has units of Ops/s and the latency, $L$, has units of seconds.

For Track 2, PDP is directly calculated from the measured power as
\begin{equation}
    PDP = P_\text{Track 2} \times L.
\end{equation}

\subsubsection{Chip resources}
The physical resource cost of mapping networks into neuromorphic architectures is an important evaluation metric since chip resources impose a hard constraint on network complexity. Compared to conventional architectures that scale through the use of bountiful off-chip memory, neuromorphic architectures embed all network configuration on-chip, hence are limited by available state for representing synaptic weights, network routing tables, neuron parameters, and other configuration parameters. 

For Loihi 2 and similar architectures, the ultimate measure of a workload's chip resource cost is {\em core count}.  For Track 2, this is the definitive chip resource utilization metric used in this challenge.

Before networks are successfully mapped to chip, it is difficult to reliably estimate core count requirements, so for Track 1, we assess solutions by indirect measures of resource cost: parameter count and total model size.

A network's parameter count includes its total synaptic state ({\eg}, weights and delays) and neuron parameters such as decay factors.  Only unique parameters are to be counted, as expected to be uniquely configured in on-chip memories and tables leveraging convolutional and other network compression features. Note that a network's {\em trainable} parameters will be a subset of its total unique configuration parameters.  

Model size is the sum over the bit widths of all unique parameters, measured in bytes.  Since Loihi 2 supports a range of synaptic weights from one to eight bits, it is possible for two networks with the same parameter counts to have very different model sizes. All else being equal, solutions with smaller model sizes are preferred.

\section{Baseline solution}\label{sec:baseline-solutions}
We provide a baseline solution for Track 1 of the \intelndnschallenge{}, available in the \href{\ndnsrepourl}{\intelndnschallenge{} Github Repository}. In this section, we outline the baseline solution architecture, a sigma-delta neural network, and the evaluation of the baseline solution on the metrics defined in \secrefprefix{} \ref{sec:evaluation}. 
Later in the challenge, we will provide a Loihi 2 version of the baseline solution and evaluate it on a Loihi 2 system; we will also release the Track 2 baseline associated code.

\subsection{Sigma-delta neural network architecture}

The proposed neuromorphic solution is a simple feedforward sigma-delta ReLU neural network (SDNN). The solution makes use of two neuromorphic computation ideologies: \emph{sparse message passing} using sigma-delta neuron and \emph{temporal computation} using axonal delays.

The delta encoding exploits the temporal similarity in the data. It sparsifies the data communicated to the next layer by sending only a change that is higher in magnitude than a certain threshold. The sigma encoding, on the other hand, reconstructs the original signal at the receiving end. A combination of sigma and delta units wrapped around a dynamics or a non-linearity (ReLU in this case) is a sigma-delta neuron \cite{oconnor2017sigma}. Sigma-delta neurons make use of the sparse messaging paradigm in neuromorphic hardware and result in a significant reduction in synaptic computations.

The axonal delays endow the network with a short-term memory capability that allows the interaction of audio/features originating at different points in time. Learnable axonal delays have been shown to increase the expressivity and performance of networks, particularly for applications with spatio-temporal features \cite{shrestha2018slayer, shrestha2022spikemax}. Audio denoising is one such application.

\begin{figure*}[!ht]
    \centering
    \includegraphics[width=0.8\textwidth]{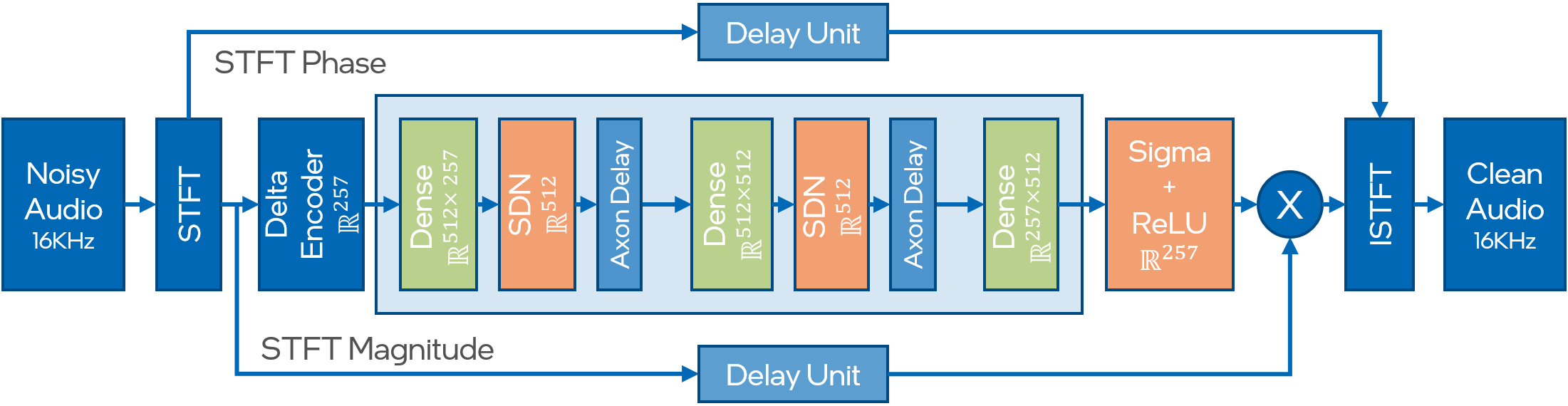}
    \caption{Sigma-delta neural network baseline solution structure. }
    \label{fig:baseline_solution}
\end{figure*}

The structure of the SDNN baseline solution is illustrated in \figrefprefix\ref{fig:baseline_solution}, and we describe the solution in the following.

\textbf{Encoder:}
The encoder is a straightforward Short-Time Fourier Transform (STFT) \cite{grochenig2001foundations} of the noisy audio waveform followed by delta encoding of the STFT magnitude. The STFT uses a \textit{window length} of 512 with a \textit{hop length} of 128 ($^1\!/\!_4$ \textit{window length}), leading to 8ms per time-step, as the signal is at 16kHz. These parameters are user-configurable. The delta encoding sparsifies the STFT magnitude which is then fed to the N-DNS network.

\textbf{N-DNS:}
The neuromorphic denoiser (N-DNS) network is a three-layer feedforward sigma-delta ReLU network with axonal delays. The sigma-delta layer efficiently performs denoising in the sparse domain. The axonal delays provide the network with short term memory which can be used to incorporate previous temporal patterns during denoising. The N-DNS network predicts a multiplicative mask at some delay which is then used to mask the STFT magnitude. The STFT phase and magnitude from the encoder need to be delayed accordingly during the decoding phase.

\textbf{Decoder:}
The decoder combines the multiplicative mask predicted by the N-DNS network with the delayed STFT phase and magnitude of the noisy audio signal and performs inverse STFT with the same \textit{window length} and \textit{hop length} as the encoder. The resulting output is the clean reconstruction (denoised) audio waveform.

The SDNN baseline network was trained with Lava-dl\footnote{Lava-dl is a deep spiking neural network training library available publicly at \url{https://github.com/lava-nc/lava-dl}}, which includes the extended version of the SNN backpropagation training tool SLAYER \cite{shrestha2018slayer}. Lava-dl SLAYER uses a surrogate gradient method ({\eg}, see \cite{neftci2019surrogate}) to address the critical challenge in training spiking neural networks---the non-differentiability of spikes. The baseline network was trained with Loihi 2's fixed precision computation in mind and trained with appropriate quantization for synapse and neuron dynamics.
We used a combination of negative SI-SNR and a mean-square error measuring the STFT magnitude reconstruction quality as the minimization loss and a RADAM optimizer for training. The detailed training procedure, as well as Lava\footnote{Lava is a neuromorphic software framework available publicly at \url{https://github.com/lava-nc/lava}} evaluation of the baseline network, are available in \href{\ndnsrepourl}{\intelndnschallenge{} Github Repository}.

\subsection{Evaluation Metrics}

We evaluated the SDNN baseline solution, Microsoft NsNet2 (the baseline network for Microsoft DNS 2022), and \ipgnet\/ using Track 1 metrics on the validation set.
The metrics are summarized in \tablerefprefix{} \ref{tab:evaluation_metrics_comparison}.
All three networks use STFT encoding and ISTFT decoding.

\ipgnet{} is an Intel proprietary network used in production. The model is causal, operates in real-time, and is built from LSTM and 2D convolution layers. Power metrics for this network are not available. The network was trained using proprietary datasets and augmentation techniques, and as such we view its audio quality results as upper-bound aspirational targets for challenge submissions.  

The audio quality metrics include DNSMOS scores, SI-SNR, and improvement in SI-SNR (SI-SNRi). The encoder and decoder for all three networks perform lossless transformation using STFT and ISTFT. As a result, relative performance differences across models in SI-SNR and SI-SNRi are equal.

The latency was calculated by summing data buffer latency, encoder-decoder latency, and network latency (N-DNS latency), as described in \secrefprefix{} \ref{sec:evaluation}. 

Power proxy and PDP proxy metrics provide some measure of the relative power and power-delay-product across the three networks suitable for Track 1 comparisons. For the SDNN baseline, these are calculated according to {\equationsprefix} (\ref{eq:power-proxy}) and (\ref{eq:pdp_proxy}), respectively. For the conventional Microsoft NsNet2 network, Ops refer to Multiply–accumulate operations (MACs) without considering the negligible cost of per-neuron ReLU evaluation.

\begin{table*}[!ht]
    \centering
    \scriptsize
    \caption{Evaluation metrics comparison.}
    \label{tab:evaluation_metrics_comparison}
    \setlength\tabcolsep{3pt} 
    \begin{tabular}{l|r|r|r|r|r|r|r|r|r|r|r|r}
         \multicolumn{1}{@{}c@{}|}{\multirow{3}{*}{\bf Network}} &
         \multirow{2}{*}{\bf SI-SNR} &
         \multicolumn{2}{@{}c@{}|}{\bf SI-SNRi} &
         \multicolumn{3}{@{}c@{}|}{\bf DNSMOS$^\ddag$} &
         \multicolumn{2}{@{}c@{}|}{\bf Latency} &
         \multicolumn{1}{c|}{\bf Power} &
         \multicolumn{1}{c|}{\bf PDP} &
         \multicolumn{1}{c|}{\bf Param} &
         \multicolumn{1}{c}{\bf Model}\\ \cline{3-9}
         & &data & enc+dec & OVRL & SIG & BAK & enc+dec$^\dagger$
         & \multicolumn{1}{c|}{total} &
         \multicolumn{1}{c|}{\bf proxy} &
         \multicolumn{1}{c|}{\bf proxy} & 
         \multicolumn{1}{c|}{\bf count} &
         \multicolumn{1}{c}{\bf size} \\
         & \multicolumn{1}{c|}{dB}
         & \multicolumn{1}{c|}{dB}
         & \multicolumn{1}{c|}{dB}
         & \multicolumn{1}{c|}{}
         & \multicolumn{1}{c|}{}
         & \multicolumn{1}{c|}{}
         & \multicolumn{1}{c|}{ms}
         & \multicolumn{1}{c|}{ms}
         & \multicolumn{1}{c|}{M-Ops/s}
         & \multicolumn{1}{c|}{M-Ops}
         & \multicolumn{1}{c|}{$\times 10^3$}
         & \multicolumn{1}{c}{KB}\\ \hline\hline
         Microsoft NsNet2    & 11.89 & 4.26 & 4.26 & 2.95 & 3.27 & 3.94 & 0.024 & 20.024 & 136.13 & 2.72 & 2,681 & 10,500\\
         \ipgnet             & 12.71 & 5.09 & 5.09 & 3.09 & 3.35 & 4.08 & 0.036 &  32.036 &  - & - &   1,901 & 3,802\\ 
         {\bf SDNN baseline} & 12.50 & 4.88 & 4.88 & 2.71 & 3.21 & 3.46 & 0.036 &  32.036 &  14.54 & 0.44 &   525 & 465\\
         Validation set (noisy)     &  7.62 &    - &    - & 2.45 & 3.19 & 2.72 & - & - & - & - & - & -\\
         \hline
         & \multicolumn{6}{c|}{higher is better ($\uparrow$)} & \multicolumn{6}{c}{lower is better ($\downarrow$)} \\
         \multicolumn{12}{p{12cm}}{\tiny$^\dagger$ Latency results measured on a system with Intel(R) Xeon(R) Platinum 8280 CPU @ 2.70GHz and 32 GB RAM as of Feb 2023 and may not reflect all publicly available security updates. Results may vary.} \\
         \multicolumn{12}{p{12cm}}{\tiny$^\ddag$ Please note that the DNSMOS scores in this table are not directly comparable to the DNSMOS scores presented in the results of the Microsoft DNS Challenge due to differing composition of validation/test sets.}
    \end{tabular}\label{table:baseline-metrics}
\end{table*}

We see that our SDNN baseline is a promising neuromorphic solution to the audio denoising problem. In terms of audio quality, the SDNN baseline has a higher SI-SNR relative to the NsNet2 baseline solution from the Microsoft DNS Challenge 2022, and lower relative DNSMOS scores. Notably, our baseline solution training targeted an SI-SNR loss, thus better relative SI-SNR performance may be expected. Nonetheless, it is encouraging to see substantial DNSMOS improvement over the unprocessed noisy input in a system not trained specifically for perceptual quality. And importantly, the SDNN solution is an order of magnitude more efficient than the NsNet2 baseline in terms of the power proxy even though it processes data at a throughput $1.25 \times$ higher than the NsNet2 baseline, and it uses $5\times$ fewer parameters. The quantization-aware training of the baseline SDNN solution further reduces the model size by $22\times$ compared to NsNet2.

Naturally, the NsNet2 solution is a baseline and does not represent state-of-the-art for audio denoising today. For example, the Intel production DNS model (Intel DNS network) achieves higher SI-SNR and DNSMOS than both NsNet2 and the SDNN baseline solution (\tablerefprefix{} \ref{tab:evaluation_metrics_comparison}). 
Given the simplicity of our SDNN baseline solution as a starting point for neuromorphic audio denoising, we believe it will be possible to significantly improve its denoising quality while also reducing its computational resources with further algorithmic innovations in the \intelndnschallenge{}.

Notably, the sigma-delta approach in our baseline solution is quite general. Sigma-delta sparsification can be applied to any conventional ReLU-like nonlinearity as well as to the dynamics present in typical neuromorphic neuron models such as leaky integrators and resonators.  
Furthermore, sigma-delta sparsification represents just one neuromorphic feature available of many to exploit by participants in the challenge. We see a wide space of uncharted waters to explore for the \intelndnschallenge{}. Our baseline solution represents just a first step, and we find it encouraging that it already provides promising results.

\section{Additional information}\label{sec:clarifications}

Please see the \href{\ndnsrepourl}{\intelndnschallenge{} Github Repository} for the official competition rules, timeline, registration procedure, metrics boards, code, and datasets. Any additional clarifications that may arise during the challenge will be posted there.

\section{Conclusion}\label{sec:conclusion}

We introduce the Intel Neuromorphic DNS Challenge to fulfill a vital need for a widely-applicable challenge problem that facilities algorithm innovation leading to a clear demonstration of neuromorphic hardware benefits. 

We include two tracks to encourage (1) algorithmic innovation and (2) demonstration on neuromorphic hardware, and we specify task performance metrics and computational cost metrics to make it easy to compare different solutions.
Furthermore, we provide permissively-licensed dataloader scripts, evaluation scripts, and an example neuromorphic baseline solution for accessibility, convenience, consistency, and extensibility. 
We also offer a monetary prize to encourage participation. 

We look forward to the learnings that we gain as a community through the \ndnschallenge{}, both in terms of the innovation that occurs in the solution space, as well as the insights that can inform the development of future neuromorphic challenge problems.

\bibliographystyle{ieeetr}
\bibliography{references}

\end{document}